\documentclass[letterpaper, 10pt,  journal, final, twocolumn]{IEEEtran}
\usepackage{graphicx}
\usepackage{multicol}
\usepackage{subfig}
\usepackage{amsmath,amssymb}
\usepackage{multirow}

\usepackage[font=small]{caption}
\usepackage{amsmath}
\usepackage{amssymb}
\usepackage{float}

\usepackage{lettrine}
\usepackage{algorithm}
\usepackage{algpseudocode}
\usepackage{cite}
\usepackage{filecontents}
\usepackage{lipsum}
\usepackage{color}
\usepackage{esdiff}
\usepackage{epstopdf}
\usepackage{flushend}
\usepackage{balance}
\usepackage[utf8]{inputenc}
\usepackage[english]{babel}
\usepackage{amsthm}
\usepackage{notoccite}

\newtheorem{defn}{Definition}

 \newcommand{\Real}{\mathbb{R}}

 \newcommand{\blue}{\color{blue}}

\begin{document}

\title{{\small\textbf{\blue IEEE/MTS Global OCEANS 2021, San Diego - Porto}}\\ \vspace{-6pt}
\Huge {CPPNet: A Coverage Path Planning Network}\\
}

\author{ \begin{tabular}{cccccccccc}
{Zongyuan Shen$^\dag$$^\star$} & {Palash Agrawal$^\dag$}  & {James P. Wilson$^\dag$} & {Ryan Harvey$^\dag$} & {Shalabh Gupta$^\dag$}
\end{tabular}\vspace{-15pt}
\thanks {$^\dag$Dept. of Electrical and Computer Engineering, University of Connecticut, Storrs, CT 06269, USA.}
\thanks {$^\star$Corresponding author (email: zongyuan.shen@uconn.edu)}
\thanks{\copyright  2021 IEEE. Personal use of this material is permitted. Permission from IEEE must be obtained for all other uses, in any current or future media, including reprinting/republishing this material for advertising or promotional purposes, creating new collective works, for resale or redistribution to servers or lists, or reuse of any copyrighted component of this work in other works.}
}

\maketitle

\begin{abstract}
This paper presents a deep-learning based CPP algorithm, called Coverage Path Planning Network (CPPNet). CPPNet is built using a convolutional neural network (CNN) whose input is a graph-based representation of the occupancy grid map while its output is an edge probability heat graph, where the value of each edge is the probability of belonging to the optimal TSP tour. Finally, a greedy search is used to select the final optimized tour. CPPNet is trained and comparatively evaluated against the TSP tour. It is shown that CPPNet provides near-optimal solutions while requiring significantly less computational time, thus enabling real-time coverage path planning in partially unknown and dynamic environments.
\end{abstract}

\IEEEpeerreviewmaketitle

\thispagestyle{empty}

\vspace{-9pt}
\section{Introduction}
Coverage Path Planning (CPP) aims to find a trajectory for an autonomous vehicle that enables it to pass over all points in the search area while avoiding obstacles\cite{song2018,SG19}. Its applications for autonomous underwater vehicles (AUVs) include seabed mapping\cite{galceran2015coverage,shen2016,shen2017,palomeras2018autonomous,paull2018probabilistic,shen2019,shen2020,shen2020ct,yordanova2020coverage}, structural inspection\cite{englot2013,vidal2017online,bircher2018receding,song2020online}, oil spill cleaning\cite{song2013,song2015slam}, mine hunting\cite{mukherjee2011}, and other underwater tasks\cite{Song2019_TStar,Wilson2020_T*Lite}. A variety of methods have been developed to solve the coverage path planning problem; a review of existing coverage methods is presented in\cite{galceran2013}. In general, coverage path planning methods can be categorized into \textit{a priori} known and \textit{a priori} unknown environments. While methods for covering unknown environments compute the paths \textit{in situ}, and are thus adaptive to the environment, methods for known environments attempt to generate the solution that optimizes a specific cost function, e.g., trajectory length, number of turns, and energy.

CPP in known environments is an application of the traveling salesman problem (TSP), in that the optimal CPP trajectory is also an optimal TSP tour. Finding the optimal solution, however, is NP-Hard. While many heuristics exist, these methods are either: i) sub-optimal, or ii) not suitable for online computation \cite{Laporte1992_TSP}. 
Many AUV missions are dynamic in nature, requiring fast online coverage path replanning when changes in the environment or mission occur. As such, it is of critical importance to develop a coverage path planner that can quickly find near-optimal tours under dynamic conditions \cite{Zeng2015_AUV}.


Many algorithms have been developed for the CPP problem in \textit{a priori} known environments. 
Zelinsky et. al\cite{zelinsky1993} proposed a grid-based method where a distance field is constructed given specific start and goal cells. The coverage path is then generated along the steepest ascent from the start to the goal.
Acar et al.\cite{acar2002morse} developed an offline method which identifies the critical points on the boundaries of obstacles for dividing the area into subregions and creating an adjacency graph. Then, the sequence that visits each node exactly once is computed by the depth-first-search method, while the coverage is achieved via back-and-forth motion in each cell. These methods can achieve complete coverage, however, the resulting coverage path may not be globally optimal. 



Huang et al.~\cite{huang2001} utilized dynamic programming to minimize the total number of turns by allowing different sweep directions in each subregion. Xie et al.\cite{xie2020path} proposed a similar approach to cover multiple disjoint obstacle-free subregions for Unmanned Aerial Vehicles (UAVs). The search space is divided into a set of subregions. Each subregion has four potential back-and-forth coverage paths with starting point in different corners. Then, a Generalized Travelling Salesman Problem (GTSP) is formulated to find the optimal subregion traversal sequence and the associated coverage paths using dynamic programming. This problem is computationally expensive, thus it is not real-time implementable in arbitrary environments; limiting real-world applications. 




\begin{figure*}[t]
    \centering
    \includegraphics[width=0.95\textwidth]{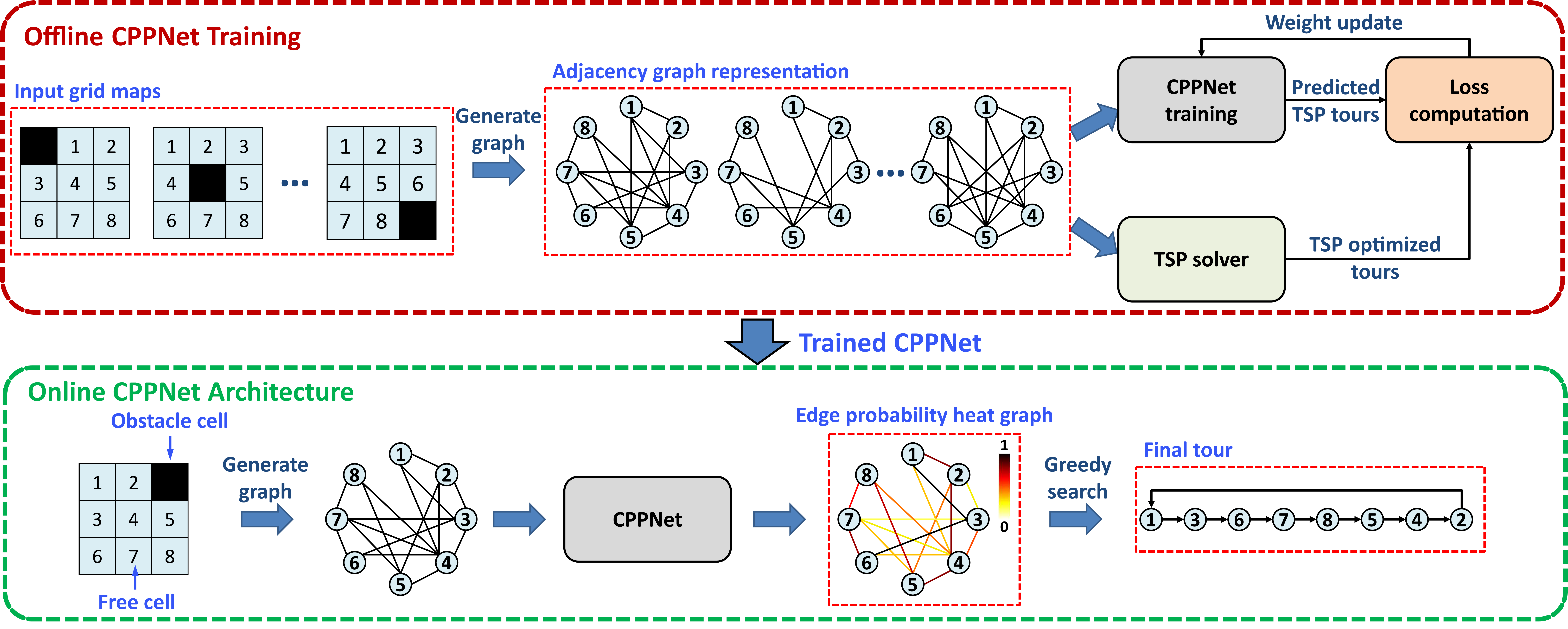}
    \caption{The training and online architecture of CPPNet.}
  \label{fig:DNN} 
\end{figure*}

Recently, many methods have been developed to utilize machine learning techniques to solve the related motion planning problems in high-dimensional spaces \cite{qureshi2020motion,wang2020neural}. These methods can find near-optimal solutions in complex scenarios much faster than the traditional approaches. 
Therefore, we developed a deep learning-based algorithm, called Coverage Path Planning Network (CPPNet), which can quickly find the near-optimal coverage path in complex environments.

CPPNet is built using a convolutional neural network (CNN), whose input is a graph-based representation of the occupancy grid map, and whose output is an edge probability heat graph, where the value of each edge is the probability of belonging to the optimal TSP tour.
A greedy search can then be performed over this heat graph to rapidly produce a near-optimal TSP tour in an environment of arbitrary complexity. CPPNet can be invoked on-demand to generate a near-optimal coverage path for the  environment with the most up-to-date information. The advantages of CPPNet are as follows: it i) produces near-optimal coverage trajectories, ii) efficiently works in environments with complex shapes, and iii) is computationally efficient for real-time implementation.

The rest of the paper is organized as follows. Section~\ref{sec:probstatement} formulates the CPP problem. Next, the CPPNet algorithm is described in Section~\ref{sec:cppnet}. The results including the generated coverage paths for some scenarios are shown and discussed in Section~\ref{sec:results}. Finally, Section~\ref{sec:conclusion} concludes the paper and provides recommendations for future work.


\section{Problem Statement}
\label{sec:probstatement}

Let $\mathcal{A}\subset\mathbb{R}^2$ be the search area which is bounded either by a hard boundary (e.g., seamount) or by a soft boundary (e.g., sub-space of a large area). A number of obstacles are populated inside $\mathcal{A}$. First, we construct a tiling on $\mathcal{A}$ as follows.

\begin{defn}[Tiling]\label{define:tiling}
A set $\mathcal{T}=\{\tau_i \subset \mathbb{R}^2, i=1,\ldots|\mathcal{T}|$\} is called a tiling of $\mathcal{A}$ if its elements: i) have mutually exclusive interiors, i.e., $\tau^0_{i} \cap \tau^0_{j} =\emptyset, \forall i \neq j$, where $i, j \in \{1,\ldots,|\mathcal{T}|\}$, and superscript $0$ denotes the interior, and ii) form a cover of $\mathcal{A}$, i.e., $\mathcal{A} \subseteq \bigcup_{i=1}^{|\mathcal{T}|  }\tau_i$
\end{defn}

The tiling $\mathcal{T}$ is separated into obstacle ($\mathcal{T}^o$) and allowed ($\mathcal{T}^a$) regions. While the obstacle cells are occupied by obstacles, the allowed cells are desired to be covered. Let $\mathcal{A}(\mathcal{T}^a)$ denote the total area of the allowed cells in $\mathcal{T}^a \subseteq \mathcal{T}$. First, we define the feasible trajectory for complete coverage.

\begin{defn}[Feasible Trajectory]\label{define:feasibletrajectory}
Given start point $\tau_s$ of the vehicle, a sequence of cells $\pi=(\pi_1,\dots,\pi_{|\pi|})$ visited by the autonomous vehicle is called a feasible trajectory if:
\begin{itemize}
\item $\pi_1 =\tau_s$.
\item $\pi_n \in \mathcal{T}^a, \forall n \in \{1,\dots,|\pi|\}$.
\item $\pi$ forms a cover of $\mathcal{A}(\mathcal{T}^a)$, i.e.,  $\mathcal{A}(\mathcal{T}^a) \subseteq \bigcup_{n=1}^{|\pi|}\pi_n$.
\end{itemize}

\end{defn}  

Let the set of all feasible trajectories in a given environment be denoted as $\Pi=\{\pi^1,\pi^2\ldots \pi^{|\Pi|}\}$. The optimal coverage path planning problem is to find the optimal and feasible solution $\pi^{\star} \in \Pi$ that has the minimum length, i.e.,
\begin{equation}
    \pi^{\star}=\mathop{\min}_{\pi \in \Pi} L(\pi)
\end{equation}
where $L(\pi)$ indicates the length of $\pi$.

\section{CPPNet Algorithm}
\label{sec:cppnet}

This section presents the proposed deep learning based CPP algorithm, called CPPNet. The architecture of the approach is shown in Fig.~\ref{fig:DNN}. CPPNet is a deep learning based planner built using a convolutional neural network (CNN) \cite{joshi2019efficient} comprised of three parts: 1) creating the dataset of 2D obstacle environments for the training and testing of CPPNet, 2) offline training of  CPPNet, and 3) online testing of CPPNet. 

\subsection{Dataset Description}


The data set for training and testing of CPPNet was created by: 1) generating a set of random 2D scenarios with various obstacle densities, 2) selecting a collision-free coverage start point, and 3) generating the optimal TSP tours. In particular, we created a total of $1384$ $10m\times 10m$ scenarios with a random obstacle density ranging from $0\%$ to $50\%$. Then, the scenarios were partitioned into a $10 \times 10$  tiling structure  consisting of $1m \times 1m$ cells. Each tiled map has up to $100$ obstacle-free cells that need to be visited depending on the obstacles. The start point was selected as the top-left corner of the map. Finally, the optimal TSP tour was found using the $2$-opt algorithm \cite{aarts2003}. Of the $1384$ scenarios, $1024$ were used for training CPPNet, $200$ for validation, and $160$ for testing.

\begin{figure*}[t]
         \centering
    \subfloat[Scene 1: obstacle density is $10\%$; trajectory length is $90.82m$; computational time is $5.288s$.]{
        \includegraphics[width=0.3\textwidth]{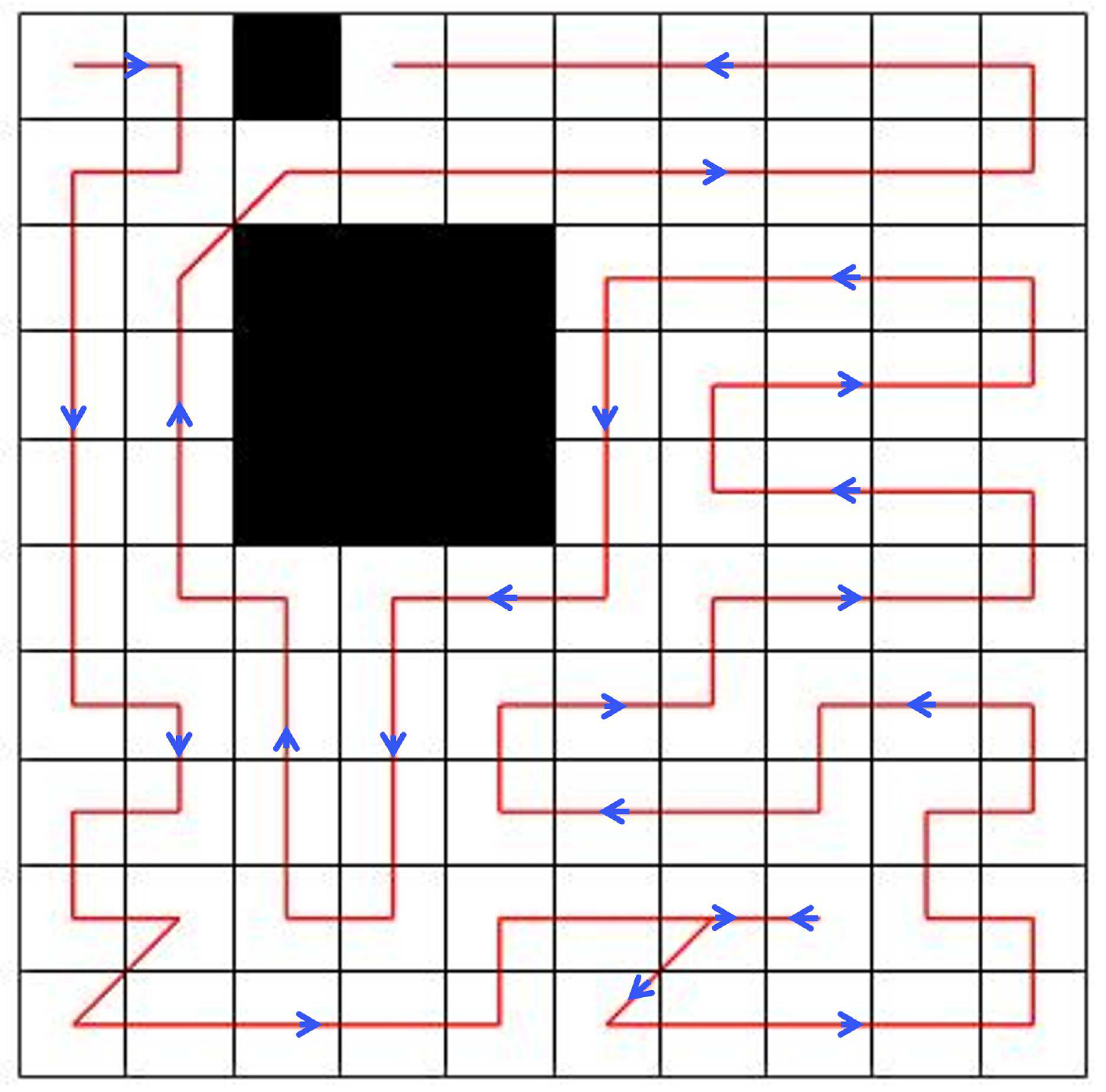}\label{fig:example_2_opt_s1}}\quad
        \centering
    \subfloat[Scene 2: obstacle density is $22\%$; trajectory length is $80.48m$; computational time is $4.636s$.]{
        \includegraphics[width=0.3\textwidth]{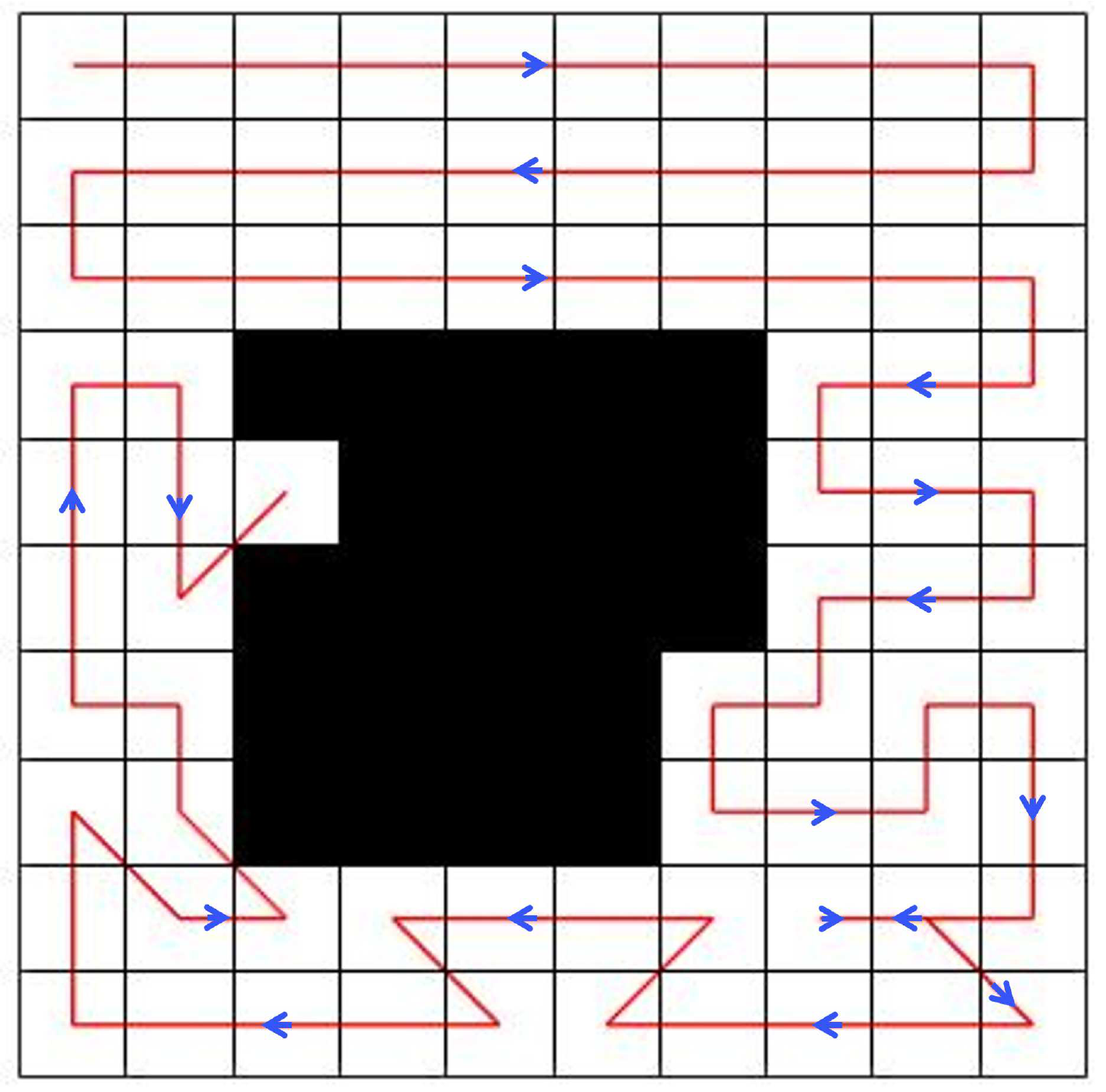}\label{fig:example_2_opt_s2}}\quad
         \centering
    \subfloat[Scene 3: obstacle density is $41\%$; trajectory length is $63.41m$; computational time is $2.999s$.]{
         \includegraphics[width=0.3\textwidth]{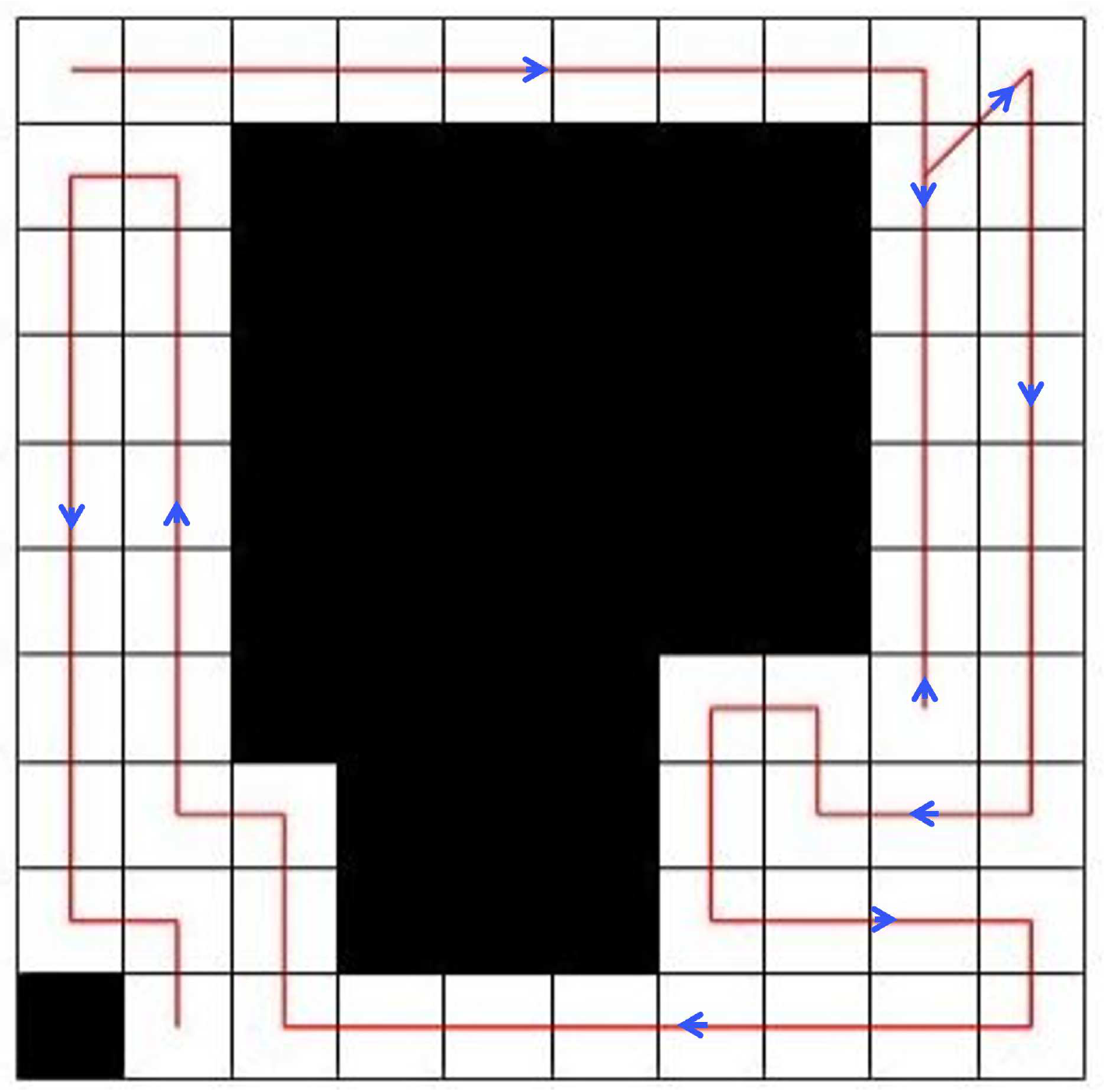}\label{fig:example_2_opt_s3}}\\
          \caption{Coverage trajectories generated by the 2-opt algorithm.}\label{fig:example_2_opt}
 \end{figure*}
 
 \begin{figure*}[t]
         \centering
    \subfloat[Scene 1: obstacle density is $10\%$; trajectory length is $92.65m$; computational time is $0.229s$.]{
        \includegraphics[width=0.3\textwidth]{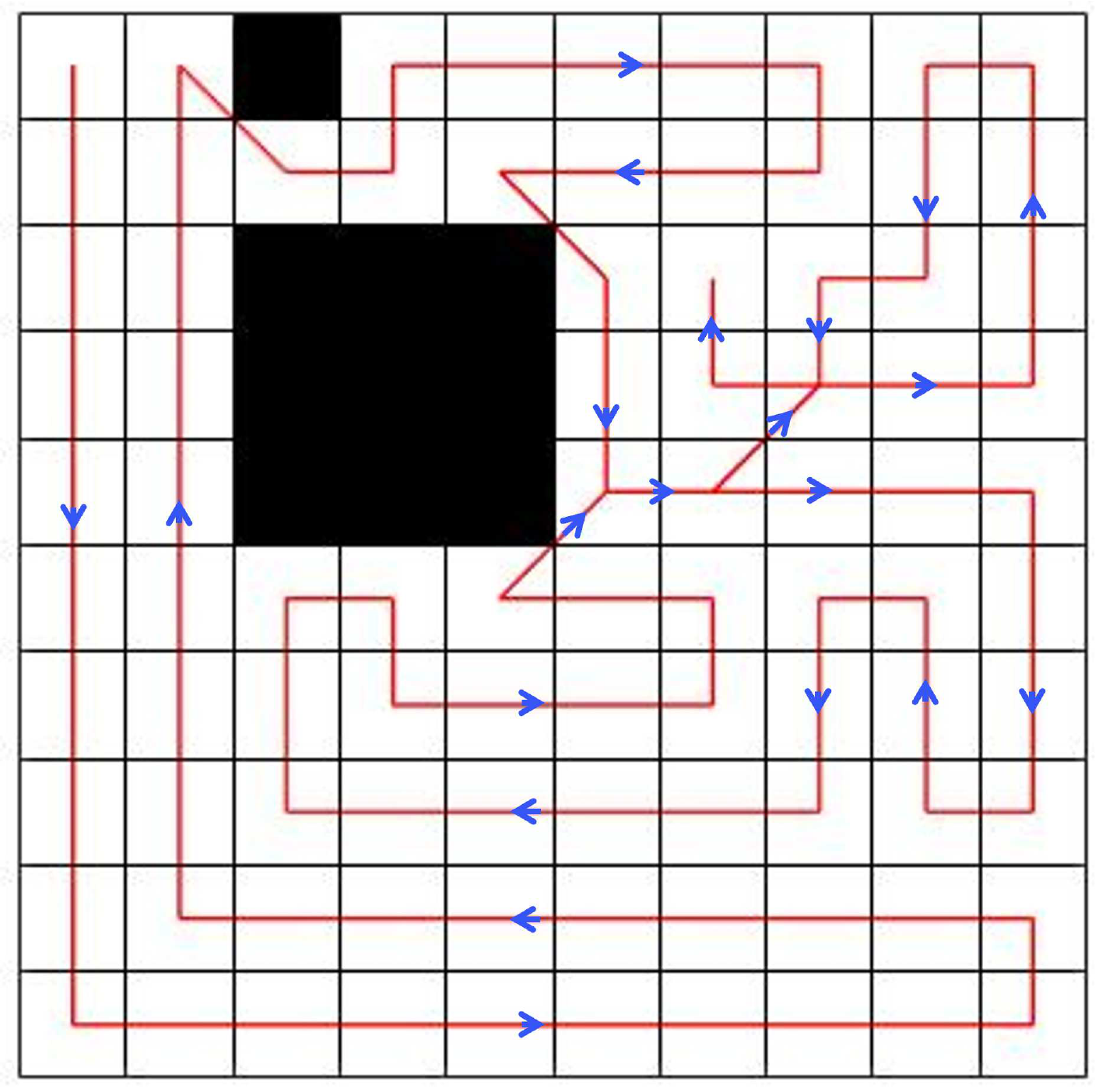}\label{fig:example_cppnet_s1}}\quad
        \centering
    \subfloat[Scene 2: obstacle density is $22\%$; trajectory length is $82.07m$; computational time is $0.176s$.]{
        \includegraphics[width=0.3\textwidth]{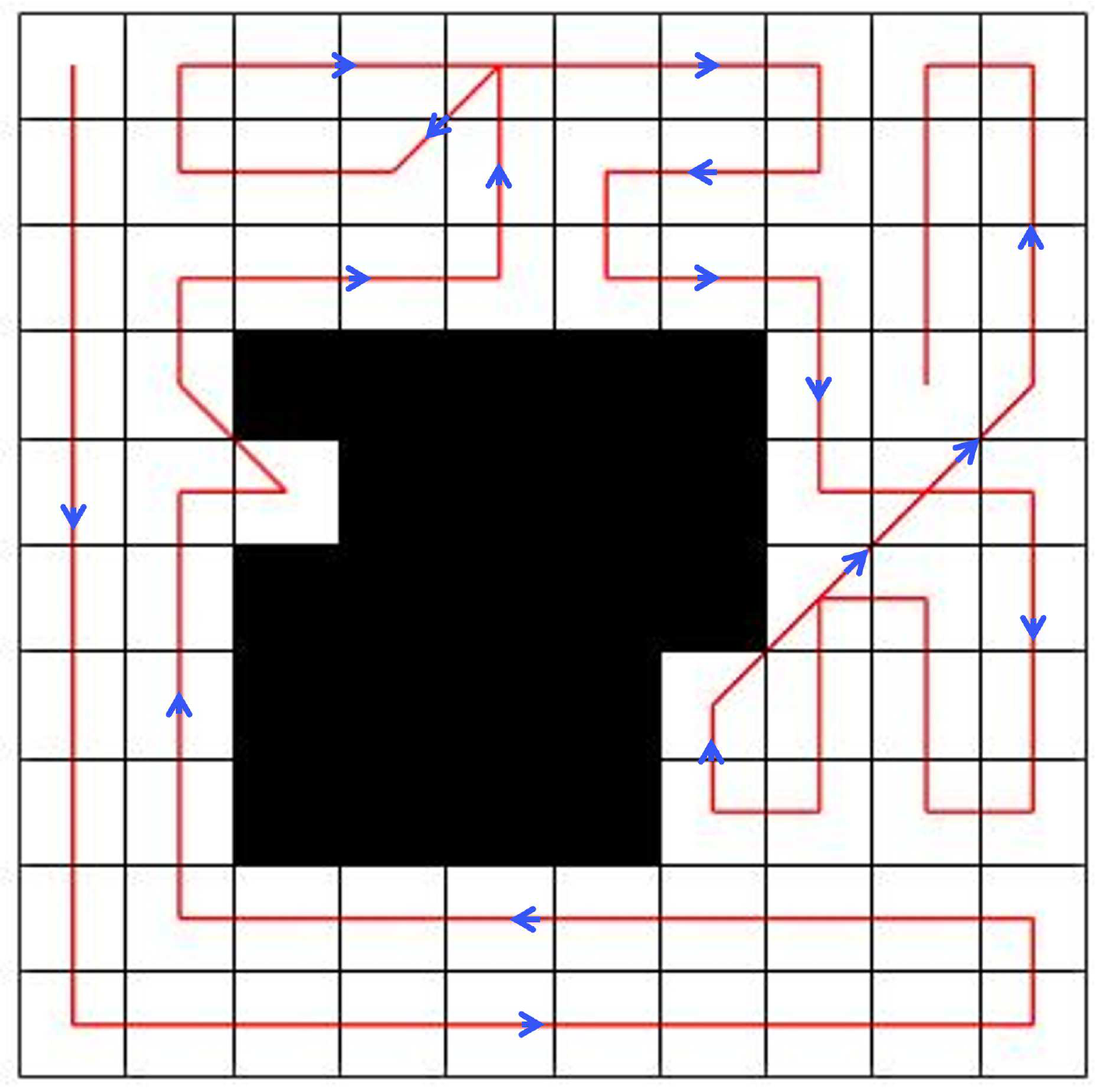}\label{fig:example_cppnet_s2}}\quad
         \centering
    \subfloat[Scene 3: obstacle density is $41\%$; trajectory length is $63.83m$; computational time is $0.162s$.]{
         \includegraphics[width=0.3\textwidth]{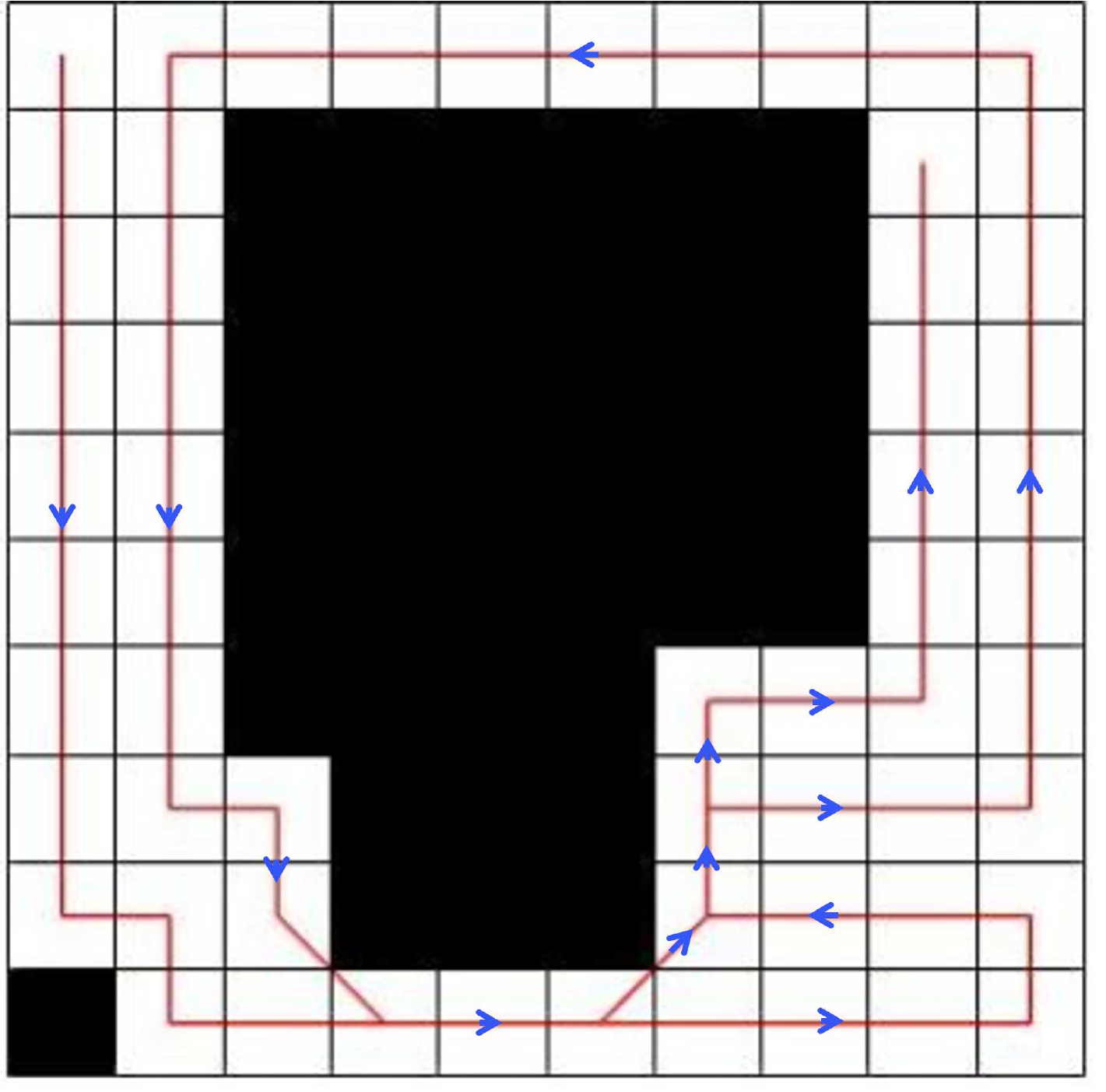}\label{fig:example_cppnet_s3}}\\
          \caption{Coverage trajectories generated by the CPPNet.}\label{fig:example_cppnet}
 \end{figure*}

\subsection{Offline Training}

CPPNet consists of three components to solve the CPP problem. The first component is the input layer which encodes the nodes and edges of the adjacency graph representation of the tiled scenario, where each node represents an obstacle-free cell and each edge indicates the connectivity between the local neighboring cells, as seen in Fig.~\ref{fig:DNN}. The second component is a sequence of Graph Convolution Layers which extracts abstract feature vectors by applying non-linear transformations to the outputs of the preceding layer. The last component is a Multi-layer Perceptron (MLP) which is used to compute the probability of an edge belonging to the optimal TSP tour. We present a summary of each of these components below; for more information, we refer the reader to \cite{joshi2019efficient}.

\vspace{6pt}
\subsubsection{Input layer}
The inputs to CPPNet are: 1) the 2D center location of each node $i$, denoted as $x_i\in\Real^2$, and 2) a $100\times 100$ adjacency matrix $E$ derived from the tiled map, which contains the distance between all node pairs $i$ and $j$, denoted as $e_{ij}$. Specifically, $e_{ij}=||x_i-x_j||$ if $i$ is adjacent to $j$ and both $i$ and $j$ are obstacle-free, and $e_{ij} = 0$ otherwise. The input layer aims to make high-dimensional feature vectors before feeding them to the Graph Convolutional Layers. The feature vector of node $i$ is \cite{joshi2019efficient}:
\begin{equation}
    x_i^0 = W^0_1x_i+b^0_1
\end{equation}
where $W^0_1\in\Real^{h\times 2}$ is a set of weights, $b^0_1\in\Real^h$ is a set of biases, $h$ is a user-defined hidden parameter, and $^0$ refers to the input layer. We also define an indicator function $\delta_{ij}$ that has a value of $1$ if nodes $i$ and $j$ are obstacle-free and adjacent, $2$ for self-connections, and $0$ otherwise. The embedding of the edge connecting node $i$ to node $j$ is:
\begin{equation}
    e_{ij}^0 = (W^0_2e_{ij}+b^0_2) \; ||\; W^0_3 \delta_{ij}
\end{equation}
where $W^0_2, W^0_3\in\Real^{\frac{h}{2}\times 1}$ are weights, $b^0_2\in\Real^{\frac{h}{2}}$ are biases, and $\cdot ||\cdot$ is the vertical concatenation operator. Note that $x_{i}^0,e_{ij}^0\in \Real^{h\times 1}$.


 \begin{figure*}[h!]
         \centering
    \subfloat[Coverage trajectory length (m).]{
        \includegraphics[width=0.475\textwidth]{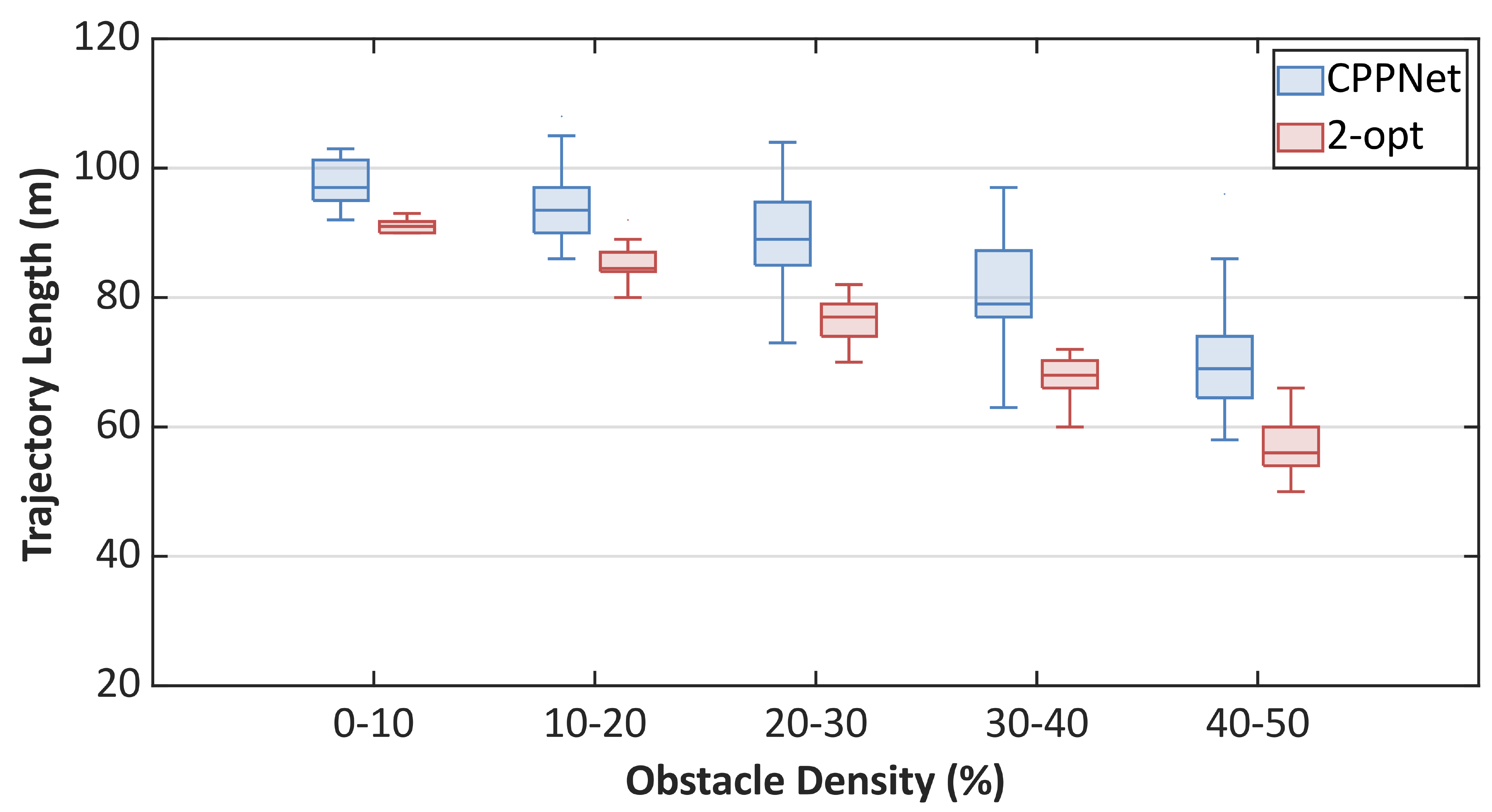}\label{fig:pathLength}}\quad
         \centering
    \subfloat[Computational time (s).]{
         \includegraphics[width=0.475\textwidth]{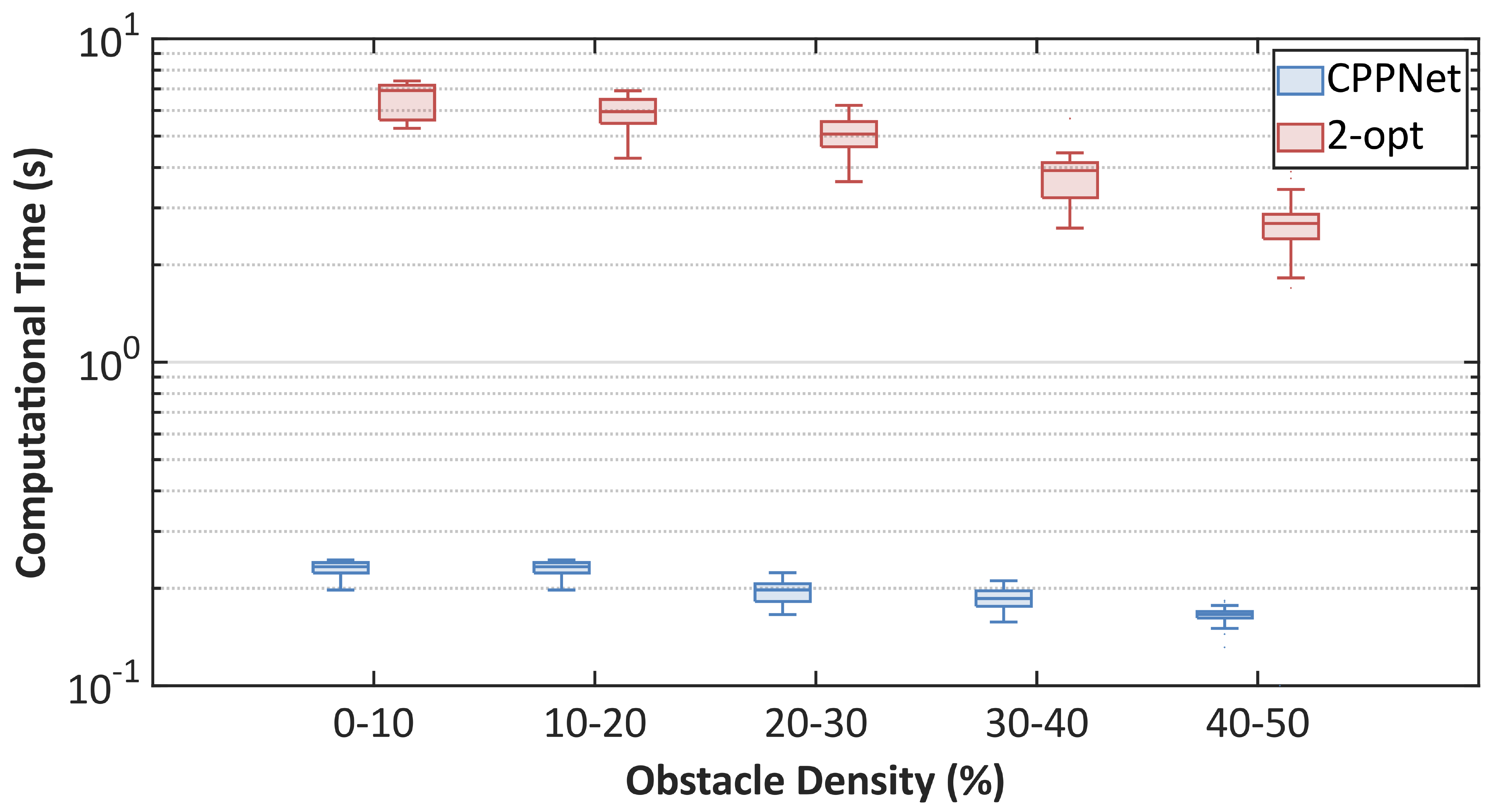}\label{fig:computeTime}}\\
          \caption{Monte Carlo simulation results.}\label{fig:MC}
 \end{figure*}

\vspace{6pt}
\subsubsection{Graph convolution layer} $x^{\ell}_i$ and $e^{\ell}_{ij}$ are the node and edge feature vectors corresponding to node $i$ and edge $ij$, respectively, at layer $\ell$. The feature vectors at the next layer are \cite{joshi2019efficient}:
\small
\begin{equation}
    x_{i}^{\ell+1}=x_{i}^{\ell}+ReLU\left \{BN\left (W_{1}^{\ell}x_{i}^{\ell}+\sum_{j\sim i}\eta _{ij}^{\ell}\odot W_{2}^{\ell}x_{j}^{\ell}  \right )  \right \}
\end{equation}

\begin{equation}
    \eta _{ij}^{\ell}=\frac{\sigma\left ( e_{ij}^{\ell} \right )}{\sum_{{j}'\sim i}\sigma\left (  e_{i{j}'}^{\ell}\right ) }
\end{equation}

\begin{equation}
    e_{ij}^{\ell+1}=e_{ij}^{\ell}+ReLU\left \{BN\left (W_{3}^{\ell}e_{ij}^{\ell}+W_{4}^{\ell}x_{i}^{\ell}+W_{5}^{\ell}x_{j}^{\ell} \right )  \right \}
\end{equation}
\normalsize
where $W^l_1,W^l_2,W^l_3,W^l_4,W^l_5\in\Real^{h\times h}$ are weight matrices, $\sigma$ is the sigmoid function, $ReLU$ is the rectified linear unit, and $BN$ is batch renormalization.

\vspace{6pt}
\subsubsection{Multi-layer Perceptron (MLP)} Each edge feature vector $e_{ij}^L$ from the last Graph Convolutional Layer $L$ is used by MLP to calculate the probability that edge $ij$ belongs to the optimal TSP tour \cite{joshi2019efficient}:
\begin{equation}
    p_{ij}^{\text{TSP}} = \text{MLP}(e_{ij}^L)
\end{equation}
After computing $p_{ij}^\text{TSP}$ for each edge $ij$, the probabilistic heat map over the adjacency matrix of connections between nodes $P^\text{TSP}$ is obtained. Thus, the output is a $100\times 100$ edge probability heat graph $P^\text{TSP}$ with probabilities $p_{ij}^\text{TSP}$, where a high probability means edge $ij$ has a high chance of belonging to the optimal solution, and a low probability means this edge likely does not belong to the optimal solution. During training, these output probabilistic heat graphs are compared against the ground truth $100\times 100$ adjacency graph $\hat{P}^\text{TSP}$ from the 2-opt solution, where $\hat{p}^\text{TSP}_{ij}=1$ means edge $ij$ belongs to the optimal solution, and 0 otherwise. The loss is then computed, and the weights of CPPNet are updated accordingly.

\vspace{6pt}
\subsubsection{Loss Function} The edge probability heat map over the adjacency graph $P^\text{TSP}$ is compared against the adjacency graph of the ground-truth 2-opt TSP solution $\hat{P}^\text{TSP}$. The weighted binary cross-entropy loss is computed and minimized using the Adam optimizer.




\subsection{Online Coverage Path Planning}

The online phase exploits the neural models from the offline training phase to perform coverage path planning in obstacle-rich environments. The overall flow of online coverage path planning is shown in Fig.~\ref{fig:DNN}. Specifically, an adjacency graph representation is generated based on a binary occupancy grid map. Then, the graph is given as the input to the trained CPPNet, which outputs a probabilistic heat graph over the adjacency matrix of tour connection. The value of each edge in the heat graph is the probability of belonging to the optimal TSP tour. A greedy search is  then used to find the final tour by selecting the next node in the local neighborhood possessing the highest probability. Note that if there is not an unvisited node in the local neighborhood, we expand the size of local neighborhood incrementally until an unvisited node is found. After the optimal TSP tour is obtained, the A$^\star$ algorithm \cite{hart1968} is used to construct the final path along this tour.

\vspace{6pt}
\section{Results and Discussion}
\label{sec:results}
In this section, the coverage paths produced by CPPNet and the 2-opt algorithm are shown and compared. CPPNet was trained on a Windows $10$ computer with an Intel Core i$5$ processor with $8$GB RAM using PyTorch. There are a total of $3$ Graph Convolutional Layers and $2$ MLP layers. The hidden parameter $h$ is $50$. The initial learning rate was set to $0.001$ while the mini-batch size was $20$. The training converged within $6$ epochs in about $16$ minutes.



Fig.~\ref{fig:example_2_opt} and Fig.~\ref{fig:example_cppnet} show the coverage trajectories of the 2-opt algorithm and CPPNet in three different scenarios, representing a low obstacle density ($0$-$10\%$), medium obstacle density ($20$-$30\%$), and high obstacle density ($40$-$50\%$), respectively. The starting point is the top-left corner of each map, and the trajectory is marked by the red color. The trajectory lengths of CPPNet are $92.65m$, $82.07m$, and $63.83m$ in the three scenarios, respectively. On the other hand, the trajectory lengths of the 2-opt algorithm are $90.82m$, $80.48m$, $63.41m$, respectively. Similar to the 2-opt trajectories, CPPNet trajectories have minimal overlap and provide high-quality solutions. The online inference times of CPPNet in these scenarios are $0.229s$, $0.176s$, and $0.162s$, respectively, while the inference times of the 2-opt algorithm are $5.288s$, $4.636s$, and $2.999s$, respectively. Thus, CPPNet generated near-optimal solutions in all three scenarios while requiring significantly less computational time as compared to  the 2-opt algorithm. 

Fig.~\ref{fig:MC} presents the box plots of the trajectory lengths (left) and online inference times (right) of the 2-opt and CPPNet solutions over all testing scenarios. In general, CPPNet is able to generate trajectories that are within $5$-$27\%$ of the 2-opt solution; however, CPPNet requires several orders of magnitude less computation time. 

Overall, CPPNet is able to produce near-optimal trajectories comparable to the 2-opt algorithm in real-time, thus making it suitable for online coverage path planning in dynamic or partially unknown environments.

\section{Conclusions and Future Work}
\label{sec:conclusion}
This paper presents a Deep Neural Network-based Coverage Path Planning algorithm, called Coverage Path Planning Network (CPPNet). CPPNet is built on a convolutional neural network (CNN) which takes the adjacency graph representation of an occupancy grid map as the input and outputs an edge probability heat graph, where the value of each edge is the probability of it belonging to the optimal TSP tour. 
A greedy search is then applied over the heat graph to produce the final TSP tour. The process concludes with A$^*$ being applied to the TSP tour to create the final coverage path. CPPNet is trained on and comparatively evaluated against the 2-opt algorithm. It is shown that CPPNet provides near-optimal solutions while requiring significantly less computational time, thus enabling online coverage path planning in dynamic environments.

In order to improve CPPNet, we first plan to create a larger and more diverse data set in order to ensure CCPNet is robust and reliable in various complex obstacle-rich environments. We will also improve the solution quality of CPPNet by including the node visitation sequence information of the near-optimal TSP solutions during training. Finally, we will develop new and scalable transfer learning approaches in order to simplify and speed up the training of CPPNet.

\balance 
\bibliographystyle{ieeetr}
\bibliography{reference}
\end{document}